\documentclass[a4paper,conference]{IEEEtran}
\ifCLASSINFOpdf
\else
\fi

\usepackage{balance} 
\usepackage{times}
\usepackage[T1]{fontenc}
\usepackage[latin9]{inputenc}
\usepackage{verbatim}
\usepackage{amsmath,amsfonts}
\usepackage{amssymb}
\usepackage{comment}
\usepackage{graphicx}
\usepackage{microtype}
\usepackage{subfig}
\usepackage{algorithm,algpseudocode}
\usepackage[inline]{enumitem}
\begin{document}
%
\title{Uncertainty Aware System Identification with Universal Policies}

\author{
\IEEEauthorblockN{Buddhika Laknath Semage*, Thommen George Karimpanal, Santu Rana and Svetha Venkatesh}
\IEEEauthorblockA{Applied Artificial Intelligence Institute\\
Deakin University\\
Geelong, Australia\\
*Email: bsemage@deakin.edu.au}}



%


\maketitle

\begin{abstract}
Sim2real transfer is primarily concerned with transferring policies trained in simulation to potentially noisy real world environments. A common problem associated with sim2real transfer is estimating the real-world environmental parameters to ground the simulated environment to. Although existing methods such as Domain Randomisation (DR) can produce robust policies by sampling from a distribution of parameters during training, there is no established method for identifying the parameters of the corresponding distribution for a given real-world setting. In this work, we propose \emph{Uncertainty-aware policy search (UncAPS)}, where we use Universal Policy Network (UPN) to store simulation-trained task-specific policies across the full range of environmental parameters and then subsequently employ robust Bayesian optimisation to craft robust policies for the given environment by combining relevant UPN policies in a DR like fashion. Such policy-driven grounding is expected to be more efficient as it estimates only task-relevant sets of parameters. Further, we also account for the estimation uncertainties in the search process to produce policies that are robust against both aleatoric and epistemic uncertainties. We empirically evaluate our approach in a range of noisy, continuous control environments, and show its improved performance compared to competing baselines.

\end{abstract}


%
\IEEEpeerreviewmaketitle

\section{Introduction\label{sec:intro}}

In recent years, deep reinforcement learning (RL) methods have achieved a number of significant accomplishments such as reaching the human level capability of playing complex games, e.g., Go \cite{silver2016mastering} and  Atari \cite{mnih2015human}. However, the sample inefficiency of these methods has been a perennial issue when applying them to real world tasks that carry high interaction costs (e.g., training time, damages caused by errors). A common approach to alleviate this issue is by using a numerical simulator that approximates real world dynamics to train RL agents before deploying them on the real-world tasks. However, precisely modelling the real-world, even in a relatively simple setting, is difficult due to the existence of dynamics that are transient or extremely expensive to model such as rolling friction and air resistance. To overcome this challenge, simulator trained policies need to be robust to these real-world noisy behaviours. 

Domain randomisation (DR) \cite{8202133,DBLP:conf/rss/SadeghiL17} is a common approach of training robust policies in the simulator, which trains an RL agent across a wide range of simulation parameters. However, one key issue with DR methods is the difficulty of determining suitable simulation parameter ranges (or distributions). In the absence of such information, the standard practice is to resort to using the full range of parameter values for DR training. Unfortunately, this results in overly robust policies that are highly suboptimal for specific instances of an environment \cite{data-driven-dr}. Thus finding a matching parameter distribution for the given environment is important. Furthermore, since all parameters may not be equally important for a given task, task-conditioned parameter learning can improve the sample efficiency of parameter matching (i.e., system identification) by learning only the parameters that the task is sensitive to. Universal Policy Networks (UPN) \cite{up-net, yu2018policy} presents an approach to obtain a library of policies trained for various combinations of simulation parameter settings. It is  particularly relevant in this regard as each trained policy, conditioned by simulation parameters, can be evaluated in the real-world to determine the best set of parameters corresponding to the task at hand \cite{lazaric2012transfer}. To expedite this policy search process, black-box optimisers such as Bayesian Optimisation (BO) \cite{frazier2018tutorial} have proven to be an effective choice \cite{yu2019sim}. However, due to the stochastic environment noise in the real-world, these methods could estimate values that are unstable and thus they are not robust to real-world conditions. 

\begin{figure}
\begin{centering}
\includegraphics[width=0.25\paperheight]{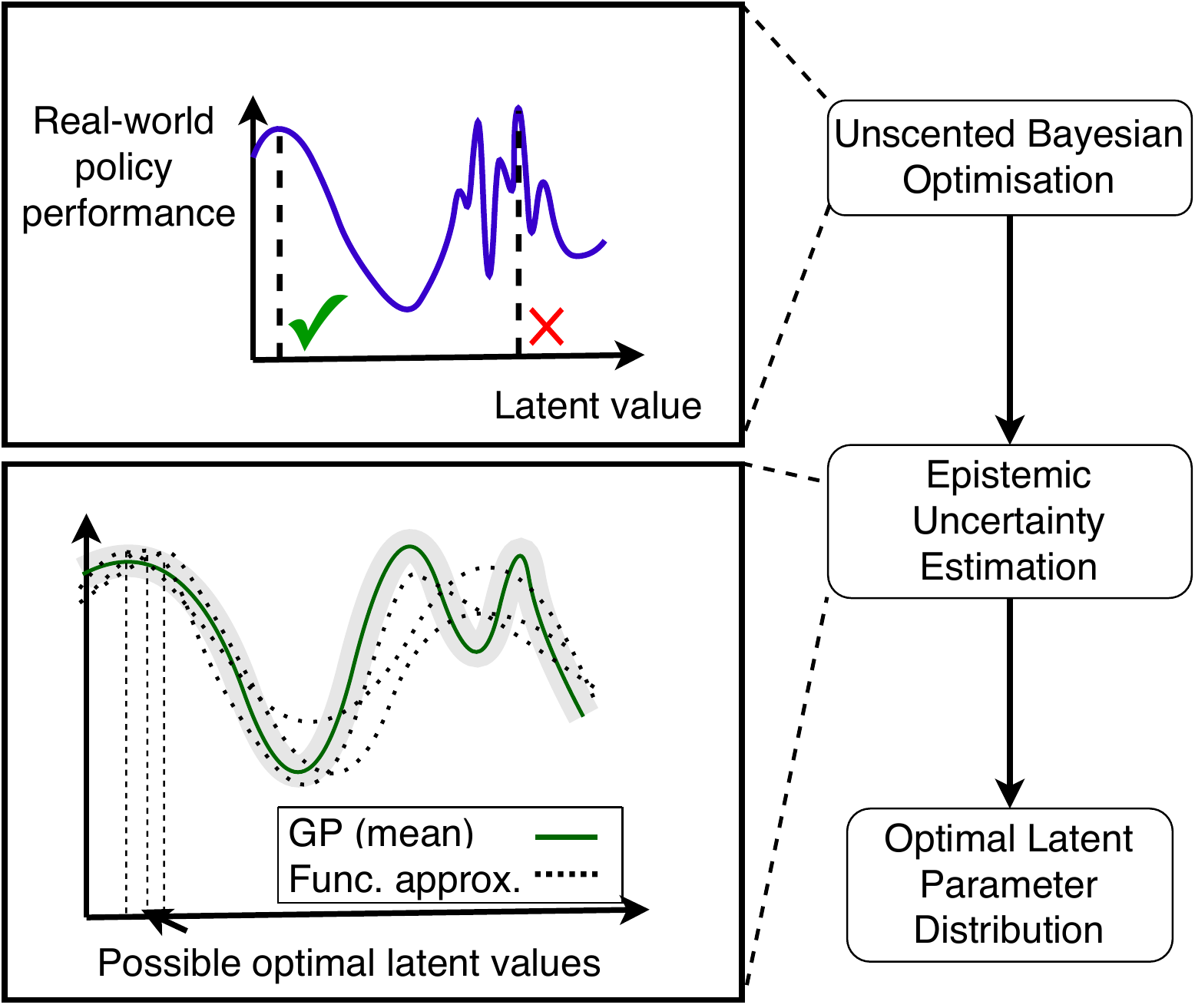}
\par\end{centering}
\centering{}\caption{\label{fig:intro} Overall workflow of \emph{Uncertainty Aware Policy Search (UncAPS)}. Aleatoric uncertainty is integrated using Unscented Bayesian Optimisation (UBO) to estimate stable latent values. Epistemic uncertainty is integrated by considering a set of function approximations of the Gaussian Process (GP) used for BO, and estimating the distribution of optimal latent values from these functions.}
\end{figure}

To address the above shortcoming, we propose a novel framework, \emph{Uncertainty Aware Policy Search (UncAPS)}, through which we aim to account for two sources of uncertainties: 1) real-world environment noise which is the cause of aleatoric uncertainty in the process, and 2) estimation error which produces epistemic uncertainty (Fig. \ref{fig:intro}). For the aleatoric uncertainty, we assume the knowledge of the variance of the noise. To incorporate a measure of aleatoric uncertainty into the policy search, we use the Unscented Bayesian Optimisation (UBO) method \cite{unscented-bo} to propagate environment noise to the parameter estimation process by using a modified acquisition function in BO, which suggests robust query points during the parameter search. Then, when devising the final robust policy from the best performing query point, we again use the unscented transformation to propagate the noise distribution to action selection by weighted averaging a set of actions given by the UPN. To address the epistemic uncertainty in policy search, instead of taking the best performing environmental parameter, we sample a set of environmental parameters from the posterior distribution of the optimal parameters. For this sampling, we follow the process as in \cite{hernandez2014predictive} that uses Random Fourier  Features \cite{rahimi2007random} as the underlying means to linearise the GP and make sampling from such a distribution tractable. We evaluate our proposed method in a range of continuous control environments and compare with the plain DR and non-robust UPN based baselines. Ablation studies show the relative importance of modelling the two types of uncertainties.

In summary, the main contributions of our study are,

\begin{enumerate}
\item Proposing a novel mechanism, UncAPS to estimate the combined aleatoric and epistemic uncertainty measures in a Universal Policy Network (UPN) based policy search workflow.
\item Devising a UPN based policy randomisation approach to integrate uncertainty measures into the policy search process to find policies that are robust to both real-world noise and estimation error. 
\item Empirically demonstrating the benefit of integrating uncertainty measures into policy search workflow on a set of continuous control environments. 
\end{enumerate}



\section{Methodology\label{sec:method}}

In this study, we propose a robust simulation-to-real system identification approach to estimate a set of unknown latent parameters in a noisy real-world environment. The underlying workflow we follow to solve this problem consists of learning a policy for each parameter setting in the simulator, and subsequently determining the simulation parameter setting of the policy that yields the best performance when evaluated in the real-world. Within this workflow, we integrate two uncertainty measures to address the estimation error and the stochasticity introduced by the noisy real-world.

\subsection{UPN Based Policy Search\label{subsec:Latent-Factor-Estimation}}

Given a set of simulation parameters, training separate policies for all combinations of parameters can be impractical, especially when the simulation parameters are continuous and high dimensional. To address this issue, we adopt an RL approach known as Universal Policy Networks (UPN) that can train policies over a given set of simulation (environmental) parameters, and interpolate policies for unseen parameters. In order to train the UPN, we consider it to be an RL agent, for which the problem is formalised as a Markov Decision Process (MDP). An MDP can be defined as a tuple of $<\mathcal{S},\mathcal{A},\mathcal{T},\theta,\mathcal{R}>$, where $\mathcal{S}$ forms the state space, $\mathcal{A}$ is the action space, and $\theta$ is the set of simulation parameters. In this setting, the transition function is given by $\mathcal{T}:\mathcal{S}\times \mathcal{A}\times\theta\rightarrow \mathcal{S}$, and reward function by $\mathcal{R}:\mathcal{S}\times \mathcal{A}\rightarrow\mathbb{R}$. UPN's state $\mathcal{S}$ consists of the combined state vectors of the task's observable state (e.g. object positions and velocity) and simulation parameters for a given episode. This formulation makes it possible for deep networks to seamlessly generalise the state, and consequently, for RL algorithms to learn policies for unseen simulation parameters.

In our problem setup, we consider two separate environments; a real world environment $\psi_r$, governed by a transition function $\mathcal{T}_r$, and a simulated (i.e., model) environment with a transition function $\mathcal{T}_s$ that we can control by changing its simulation parameters $\theta$ ($\theta \in \mathbb{R}^d$). Real-world's transition function $\mathcal{T}_r$ is parameterised by $\Theta=\theta_r + \epsilon \, (\Theta, \theta_r, \epsilon \in \mathbb{R}^d)$, where $\theta_r$ represents a fixed set of real-world latent parameters which is unknown to the RL agent, but the relationship between the latent parameters and their effects on $\mathcal{T}_r$ can be modelled universally (i.e., modellable). In contrast, $\epsilon$ represents unmodellable dynamics (i.e., environment noise) such as rolling friction and air-resistance that may not stay fixed throughout an episode, and as such, cannot be modelled accurately in the simulation. In terms of the uncertainty of modelling real-world dynamics, uncertainty of $\epsilon$ can be considered as the aleatoric uncertainty, whereas the uncertainty in estimating $\theta_r$ can be considered as the epistemic uncertainty. 

Our goal of modelling real-world dynamics in the simulation is to learn a policy $\pi$ under dynamics $\mathcal{T}_s$ that can maximise the performance when transferred to a real-world task set $W$, defined by a fixed reward function and the dynamics $\mathcal{T}_r$. The approach we take to achieve this goal is fetching $T$ number of policies $\pi_{n=1,..,T}$ from the UPN and evaluating them on $W$ to determine the best performing policy. We use Bayesian Optimisation (BO) in this process to produce a sequence of simulation parameters $\theta_{n=1,..,T}$ that will lead to a high performing policy under a relatively small budget of $T$ when $\theta_{n}$ is used to fetch a policy from the UPN. We refer to this iterative simulation parameter discovery and policy evaluation workflow as the "policy search" process.

One drawback of policy search process is its inability of exactly modelling the real-world in the simulation environment due to the stochasticity of environment noise $\epsilon$, which can lead to unstable and thus, suboptimal policies that do not generalise. Furthermore, even the estimates of modellable parameters $\theta_r$ may contain inaccuracies due to estimation error if $T$ is not large. Therefore, a policy $\pi$ trained in the simulation needs to be robust to uncertainty introduced by both of these aspects for it to consistently perform well in noisy real-world environments. For this purpose, we incorporate both aleatoric and epistemic uncertainty of modelling real-world into the policy search process.

\subsection{Integrating Aleatoric Uncertainty in Policy Search \label{subsec:ubo}}

To integrate a measure of aleatoric uncertainty into the policy search process we use Unscented BO (UBO) formulation \cite{unscented-bo}. In essence, UBO considers the input noise during the process of selecting an input point in order to select safe regions. This is achieved by selecting a set of samples deterministically from the original noise distribution (i.e., sigma points), and transforming them through a nonlinear function $f(.)$, which in our setup is a Gaussian Process (GP) used for the BO policy search. This process is referred to as unscented transformation (UT). In UT, $2d+1$ sigma points are selected as follows. If prior distribution (i.e., noise) is sampled from a Gaussian distribution $x \sim \mathcal{N}(\hat{x}, \Sigma_x)$,

\begin{align}
\begin{aligned}[t]
\label{eqn:ut}
	x^0 =& \hat{x} \\
	x_+^{(i)} =& \hat{x} + (\sqrt{(d+k)\Sigma_x})_i, \,\,\forall_i=1..d \\
	x_-^{(i)} =& \hat{x} - (\sqrt{(d+k)\Sigma_x})_i, \,\,\forall_i=1..d \\
\end{aligned}
\end{align}

where $(\sqrt{.})_i$ is the i-th row matrix square root, $k$ is a hyper-parameter that controls the scale of sigma points, and $d$ is the dimensionality of the noise sample $x$. With the samples collected, the transformed distribution is $x^\prime  \sim \mathcal{N}(\hat{x}^\prime, \Sigma_x^\prime)$ where, 

\begin{equation}
\label{eqn:ut_transform}
\hat{x}^\prime = \omega^0 f(x^0) + \sum\limits_{i=1}^{d} \omega_+^{(i)} f(x_+^{(i)}) + \omega_-^{(i)} f(x_-^{(i)})
\end{equation}

and,
\begin{align}
\begin{aligned}[t]
\label{eqn:ut_weights}
	\omega^0 =& \frac{k}{d + k} \\
	\omega_-^{(i)} = \omega_+^{(i)} =& \frac{1}{2(d + k)}, \,\,\forall_i=1..d \\
\end{aligned}
\end{align}

We integrate unscented transformation in two ways within our policy search process. 

\begin{enumerate}

\item We integrate it in the Expected Improvement (EI) acquisition function \cite{mockus1994application} used for Bayesian Optimisation by defining $f(.) \equiv EI(.)$ in Eq. \ref{eqn:ut_transform}, which is referred to as Unscented Expected Improvement (\emph{UEI}) \cite{unscented-bo}. For this purpose, to calculate the sigma points in Eq. \ref{eqn:ut}, we consider the environment noise $x \in \epsilon$ is distributed according to an isotropic multivariate normal distribution $\mathcal{N}(0, I\,\sigma^2)$.

\item To propagate environment noise to the action selection, we use UPN fetched actions to be transformed according to Eq. \ref{eqn:ut_transform} where $f(.) \equiv UPN(.)$, which we refer to as Unscented Action Selection (\emph{UAS}). When calculating sigma points in Eq. \ref{eqn:ut}, we consider simulation parameters $x \equiv \theta \sim \mathcal{N}(\hat{\theta}, I\,\sigma^2)$, where $\hat{\theta}$ is the GP suggested simulation parameter value and $\sigma^2$ is the variance of environment noise. 

\end{enumerate}

\subsection{Integrating Epistemic Uncertainty in Policy Search \label{subsec:epistemic}}

When estimating real-world parameters $\Theta$ using $\theta$ in simulator, the policy search process may carry an uncertainty introduced by the GP, especially in the early iterations of the BO process. To address this epistemic uncertainty in the policy search, we first approximate the Gaussian Process prior $f_{gp}()$ used to model the real-world policy evaluation function with a linear model $f_{gp}(\theta) = \phi(\theta)^T\hat{w}$ where $\hat{w} \sim \mathcal{N}(0, I)$ and $\phi(.)$ ($\mathbb{R}^d \to \mathbb{R}^m$) is a feature mapper for input $\theta$. With this formulation, we can sample $\hat{w}$ from the posterior of $f_{gp}$ when conditioned by observations $D_n = \{(\theta_1, y_1),..,(\theta_n, y_n)\}$ as, 

\begin{equation}
\label{eqn:gp_posterior}
\hat{w}|D_n \sim \mathcal{N}(A^{-1}\Phi^T \textbf{y}_n, \sigma_N^2A^{-1})
\end{equation}
where $A=\Phi^T\Phi + \sigma_N^2I$, $\textbf{y}_n=\{y_1,..,y_n\}$, $\Phi^T=[\phi(\theta_1)..\phi(\theta_n)]$ and $\sigma_N^2$ is observation noise variance \cite[Sec. 2.1]{hernandez2014predictive}. As for the feature mapper $\phi$, we adopt random Fourier features \cite{rahimi2007random} to map input data to a randomised low-dimensional feature space, 

\begin{equation}
\label{eqn:feat_gen}
\phi(\theta) = \sqrt{\frac{2}{m}}[cos(\omega_1^T\theta+b_1),..., cos(\omega_m^T\theta+b_m)]
\end{equation}

where $\omega_1,..,\omega_m \in \mathbb{R}^d$ are sampled from the Fourier transform $p$ of a shift-invariant kernel $k$, 

\begin{equation}
\label{eqn:fourier_feat}
p(\omega) = \frac{1}{2\pi}\int k(\delta)e^{-i\omega^T\delta}d\delta
\end{equation}

and $b_1,..,b_m \in \mathbb{R}$ are sampled from a uniform distribution on $[0, 2\pi]$.

With $N$ samples of $\hat{w}$ drawn from the posterior weights $f_{gp}$ using Eq. \ref{eqn:gp_posterior}, and corresponding $N$ feature mappers constructed using Eq. \ref{eqn:feat_gen}, we construct $N$ number of functions $f_{gp}^{(i)}(\theta)=\phi^{(i)}(\theta)^T\hat{w}^{(i)}, \forall_i={1..N}$. Then we maximise each of the $f_{gp}^{(i)}$ to find the $\theta_*^{(i)} = \underset{\theta}{argmax}\, f_{gp}^{(i)}(\theta)$. Instead of a fixed estimate of the real-world latent parameters, we use the distribution induced by the $N$ number of $\theta_*^{(i)} \in \Theta_*$ accumulated from this procedure. To apply this distribution to downstream tasks, we use the average of UPN policies fetched for each of $N$ latent parameters from $\Theta_*$ concatenated with the task's observable state.



A complete workflow of our approach is given in Algorithm \ref{alg:full}.

\begin{algorithm}[H] 
\caption{Epistemic Uncertainty Integrated Unscented Policy Search}
\label{alg:full} \hspace*{\algorithmicindent}
\textbf{Input:} \begin{algorithmic}[1]
\State $U$ - a simulator trained Universal Policy Network
\State $W$ - a set of real-world tasks
\State $\psi$, $\psi_r$ - a simulated and real-world environments
\State $T$ - Number of policy search iterations to run
\State $GP$ - A Gaussian Process model with kernel $k$
\State $\sigma^2$ - Environment noise variance
\State $N$ - Number of optimal $\theta$ samples to draw
\State $M$ - Number of Fourier random features to use

\State \textbf{Output:} Optimal simulation parameter dist. $\Theta^*$ that maximises UPN fetched policy $\pi | \Theta^*$ on real-world tasks $W$
\smallskip 

\State $D  \gets \varnothing$
\State //Run the BO policy search for T iterations
\For{$i \gets 1$ to $T$}
\State //UEI suggests the next simulation parameter that will likely improve the optimisation objective and be robust to noise. \ref{subsec:ubo} (1)
\State $\theta_i \gets UEI(GP, \sigma^2)$ 
\State // Unscented Action Selection with $\theta_i$ as noise mean in \ref{subsec:ubo} (2)
\State $\pi_i \gets$ $UAS(U, \theta_i, \sigma^2)$
\State $y_i \gets$ $\pi_i$ evaluated on task set $W$ in real-world $\psi_r$
\State $D \gets D \cup (\theta_i, y_i)$
\State $GP \gets D$
\EndFor

\State $\Theta^* \gets opt\_latent\_dist(D, N, M)$ // Optimal $\theta$ dist. using Algo. \ref{alg:epist_ubo}

\end{algorithmic} 
\end{algorithm}

\begin{algorithm}[H] 
\caption{Estimating Epistemic Uncertainty (\emph{opt\_latent\_dist})}
\label{alg:epist_ubo} \hspace*{\algorithmicindent}
\textbf{Input:} \begin{algorithmic}[1]
\State $D_n = \{(\theta_1, y_1),..,(\theta_n, y_n)\}$ - Observations from BO
\State $N$ - Number of optimal $\theta^*$ samples to draw
\State $M$ - Number of Fourier random features to use
\State \textbf{Output:} Optimal parameter distribution $\Theta^*$ 
\smallskip 
\State Initialise: $\Theta^* \gets \varnothing$

\For{$i \gets 1$ to $N$}

\For{$m \gets 1$ to $M$}
\State $\omega^{(i)}_{m} \gets$ sample $\omega$ from $p$: Eq. \ref{eqn:fourier_feat}
\State $b^{(i)}_{m} \gets$ sample from $[0, 2\pi]$
\EndFor

\State $\phi^{(i)} \gets $ Construct feature mapper with $\omega^{(i)}$ and $b^{(i)}$: Eq. \ref{eqn:feat_gen}

\State $\hat{w}^{(i)} \gets $ Sample posterior $f_{gp}$ distribution: Eq. \ref{eqn:gp_posterior}
\State Construct $f_{gp}^{(i)}(\theta) =\phi^{(i)}(\theta)^T\hat{w}^{(i)}$
\State $\Theta^*_i \gets optimise(f_{gp}^{(i)})$
\EndFor

\end{algorithmic} 
\end{algorithm}

\section{Experiments\label{sec:Experiments}}

\subsection{Experiment Setting\label{subsec:exp-setup}}

To demonstrate the general applicability of our approach, we consider a set of
five MuJoCo tasks: Half Cheetah, Hopper, Ant, Swimmer and Humanoid \cite{openai-gym}, 
implemented using MuJuCo physics simulator \cite{todorov2012mujoco}. For each of the tasks, 5 latent parameters (mass of 3 body parts, restitution, friction) are unknown to the agent (i.e., 5D). For the purpose of this study, we use two MuJoCo instances 
for each of the tasks to maintain real-world and model environments. To differentiate the real-world from the model, we add 
noise drawn from a normal distribution of mean $0$ and standard deviation $0.3$ to the real-world transition states. In practical terms this noise can be interpreted as parameters that are difficult to model universally, e.g. rolling friction, air-resistance, which causes the sim-to-real gap. Furthermore, we assume this noise distribution is known to the RL agent.


\subsection{UPN Based Policy Search with Bayesian Optimisation}

We adopt an existing UPN implemention\footnote{https://github.com/VincentYu68/policy\_transfer}\cite{yu2018policy} to train UPNs for MuJoCo tasks. 
At its core, the UPN uses an augmented state produced by concatenating task's state observations
at each step with the simulation parameters used. Using this new state, UPN uses the Proximal Policy Optimization (PPO)
algorithm \cite{DBLP:journals/corr/SchulmanWDRK17} to learn a policy. For each of the five MuJoCo tasks, we train separate UPN instances for $2\times10^{8}$ steps in the simulator.

With a given trained UPN and a MuJoCo environment with 5 unknown latent parameters, we execute 25 iterations of Bayesian optimisation (BO) to search
for a good policy by evaluating UPN trained policies on the real-world task. To provide a stable learning objective for BO, we use the cumulative reward over a window of 500 interaction steps as the optimisation objective for all tasks. For BO to depend only on the effects of simulation parameter estimates, we fix the initial state of each iteration. After the BO policy search phase, the best performing simulation parameter estimate is selected and the UPN's policy at that parameter is evaluated on the real-world task over 100 random episodes with a 500 step horizon.



\subsection{Evaluation of Uncertainty Integrated Policy Search \label{subsec:exp-eval-bo}}

To examine the effect of integrating aleatoric and epistemic uncertainty estimates in the policy search, we examine three variants of our method.

\emph{Uncertainty Aware Policy Search (UncAPS)}: To accommodate aleatoric uncertainty of the real-world, Unscented expected improvement (UEI) \cite[Section 3.3]{unscented-bo} is used as the acquisition function during policy search. The candidate simulation parameter value returned by UEI is then used as the mean value, on the basis of which an additional $2d$ simulation parameters (i.e., 10 values) are chosen. These parameter values are used to condition the UPN and fetch respective actions, which are used to calculate a weighted aggregate action as discussed in Section \ref{subsec:ubo}.
To integrate epistemic uncertainty, we approximate the GP used in \emph{UBO} with a linear model of $M=2000$ sampled Fourier Random Features (Eq. \ref{eqn:feat_gen}, \ref{eqn:fourier_feat}), and use it to calculate the posterior distribution (Eq. \ref{eqn:gp_posterior}). We repeat this for $N=250$ iterations, for each of which a set of weights is sampled from the posterior, a linear function approximation of the GP is constructed and optimised to estimate the set of optimal simulation parameters. Sampled  optimal latent parameters are then used to build an averaged policy, which we evaluate in the real-world task using the initial episodic cumulative reward of the transferred policy (i.e., jumpstart).

\emph{UncAPS$-$EP}: As an ablation study, we examine the performance of integrating only aleatoric uncertainty of the real-world, i.e., without the epistemic uncertainty estimation in \emph{UncAPS}.

\emph{UncAPS+GA}: As another ablation study, we examine \emph{UncAPS} with fitting a Gaussian distribution on optimal latent parameters sampled, and using the variance of this distribution for Unscented Action Selection (\emph{UAS}) in jumpstart evaluation. Due to using only $2d+1$ latent parameters (i.e., $11$ in our setting), this method is faster to evaluate compared to \emph{UncAPS}.



We compare our methods with the following baselines. 

\begin{itemize}
\item \emph{Standard BO}: Bayesian Optimisation policy search process using Gaussian Process (GP) as a surrogate function. We use RBF kernel and Expected improvement (EI) \cite{mockus1994application} as the acquisition function. 
\item \emph{Domain Randomisation(DR)}: simulation parameters are uniformly sampled from the same ranges used for their respective UPNs, which are then used to simulate episodes for each task. A policy is learnt on these randomised transitions in the simulator, which is then transferred and trained in the real-world to match the number of steps BO policy search used.
\item \emph{MAML}: Model agnostic meta learning (MAML) \cite{finn2017model} agents trained for each MuJoCo task in the simulator, which are then trained in the real-world (i.e., adaptation) for an equal number of steps as the BO policy search.
\end{itemize}


Table \ref{tab:mujoco-results} shows the jumpstart performances of \emph{UncAPS} against other baselines in five MuJoCo tasks. The results highlight the consistent performance gain of \emph{UncAPS} over standard BO. In three tasks (Ant, Humanoid, Half-Cheetah), epistemic uncertainty integration performs significantly (statistically) better than only considering aleatoric uncertainty (i.e., \emph{UncAPS$-$EP}), whereas in Swimmer and Hopper, the performance is within the error threshold. \emph{UncAPS+GA}, which is faster than \emph{UncAPS} to evaluate in jumpstart, performs equally or better than standard BO, but does not match to \emph{UncAPS} in Humanoid and Hopper, likely due to the optimal latent parameter distribution not conforming to a Gaussian. However, it could be a viable alternative where fast evaluations in the real-world are required. The DR baseline, which we can consider as an upper bound of the robustness, performs significantly better in Half-Cheetah, but performs considerably poorer in all other tasks compared to BO methods, likely due to conservative nature of the corresponding policies. In Half-cheetah, we hypothesise the superior performance of DR is due to a coincidental low randomisation in task dependent simulation parameters, thus avoiding a conservative policy. We also observed that \emph{MAML} baseline performs significantly poorer than the rest of the baselines considered in our setting. However, we noticed \emph{MAML}'s performance increasing with more adaptation samples being made available than the interaction budget used in our setting. It is apparent from the results that MAML may not be suitable in such ultra-low sample regimes where our BO-based method can still provide decent performance. Overall, these results underscore the benefits of integrating two uncertainty estimates in BO policy search over standard BO, unconstrained randomisation of DR and traditional meta-learning.

\begin{table*}
\centering{}%
\begin{tabular}{|c|c|c|c|c|c|c|}
\hline 
 & \emph{MAML} & DR & BO & \emph{UncAPS$-$EP} & \emph{UncAPS+GA} & \emph{UncAPS}\tabularnewline
\hline 
\hline 
Ant & 296.49 $\pm$ 0.74 & 1668.79 $\pm$ 15.90 & 1819.31 $\pm$ 9.60 & 1860.90 $\pm$ 8.29 & \textbf{1943.70 $\pm$ 8.45} & \textbf{1949.04 $\pm$ 9.33}\tabularnewline
\hline 
Humanoid & N/A & 3243.33 $\pm$ 15.13 & 3268.29 $\pm$ 3.95 & 3281.34 $\pm$ 4.40 & 3277.51 $\pm$ 4.10 & \textbf{3291.34 $\pm$ 4.92}\tabularnewline
\hline 
Swimmer & 20.79 $\pm$ 0.12 & 41.70 $\pm$ 0.25 & 61.68 $\pm$ 0.23 & \textbf{62.77 $\pm$ 0.21} & \textbf{62.40 $\pm$ 0.19} & \textbf{62.45 $\pm$ 0.20}\tabularnewline
\hline 
Half-Cheetah & 389.54 $\pm$ 1.94 & \textbf{1646.11 $\pm$ 6.24} & 1507.44 $\pm$ 4.34 & 1555.24 $\pm$ 4.09 & 1573.84 $\pm$ 4.46 & 1569.42 $\pm$ 4.78\tabularnewline
\hline 
Hopper & 930.50 $\pm$ 2.88 & 1205.17 $\pm$ 4.16 & 1277.78 $\pm$ 2.65 & \textbf{1287.50 $\pm$ 2.65} & 1276.59 $\pm$ 2.67 & \textbf{1283.49 $\pm$ 2.72}\tabularnewline
\hline 
\end{tabular}\caption{\label{tab:mujoco-results} Initial cumulative episodic reward (jumpstart) of \emph{UncAPS} against other BO variants, Domain Randomisation
(DR) and \emph{MAML} agents for five MuJoCo tasks. For all tasks, a 500 step horizon is used and jumpstart is averaged over 100 random episodes.
Bold indicates the best performance.}
\end{table*}

\section{Related Work\label{sec:bg}}

Simulation based transfer (i.e., Sim2Real) is based on discovering a shared basis between the real-world and the simulated model to successfully apply (i.e., transfer) policies learnt in the model to real-world applications. A straightforward approach of finding such a basis is matching real-world latent parameters of objects (e.g. friction, mass) and environment (e.g. wind speed) in simulation (i.e., system identification) by minimising the trajectory difference of objects in real-world and model \cite{using-inaccurate-models,farchy2013humanoid,10.5555/3304889.3305112,DBLP:conf/icra/ChebotarHMMIRF19,du2021auto, gat2017, stocgat20, allevato2020tunenet, Allevato:2020ui}. However, these approaches suffer from few key issues; a) a given task may not need to estimate all simulation parameters, where doing so would be a waste of resources (e.g. time, computational cost); b) finding a measure of trajectory difference that guarantees to estimate the true latent parameters is challenging \cite{DBLP:conf/icra/ChebotarHMMIRF19}, and c) iterative simulation parameter estimation and policy training is impractical for many real-world applications that need fast responses.

Direct policy search based methods \cite{evalutionary-direct-search} address some of these issues by evaluating simulator trained policies on real-world task sets and estimating task-conditioned simulation parameters that achieve the best transfer performance for a given task. Since the estimated parameters are task-conditioned, the search process only estimates parameters that are required for the task. However, iteratively training policies for different simulation parameters in this approach is impractical for time sensitive tasks, which highlights the need for a library of trained policies. Universal Policy Networks (UPNs) \cite{up-net} address this requirement by training a deep RL agent using an augmented state, in which the observation is combined with the true simulation parameters. When evaluating policies, this agent acts as a large collection of policies where each policy can be retrieved by conditioning with a simulation parameter. Furthermore, UPNs have demonstrated the ability to generalise its policy learning for unseen simulation parameters during training, which makes it a good solution for direct policy search based parameter estimation \cite{yu2019sim,yu2018policy}. Although these methods address a number of issues with system identification, whether the estimations given are robust to environment noise (e.g. wind, rolling friction) has not been addressed.

Domain randomisation (DR), which trains RL agents using a collection of randomised simulation parameters, is a standard approach for generating robust policies \cite{8202133,DBLP:conf/rss/SadeghiL17,matas2018sim}. Unlike with UPNs, the agent is unaware of the actual simulation parameter used in DR training, which gives rise to policies that are invariant to parameter differences. Even though DR has demonstrated impressive results in solving Rubik's Cube with a robot hand \cite{akkaya2019solving}, one major issue with the technique is determining appropriate parameter ranges to train the agent on, where using large parameter ranges have shown to produce conservative and overly robust policies \cite{data-driven-dr}. In contrast, over randomising a DR agent during training has also shown to have negative effects \cite{matas2018sim}. These studies highlight the applicability of DR to generate robust policies, but determining appropriate parameter ranges is essential for the transfer performance. 

To calibrate the parameter ranges for DR, Ramos et al. \cite{BayesSimRamosPF19} uses a likelihood-free inference method to approximate the posterior distribution of simulation parameters when the real-world object trajectories are observed. Since they use goal-independent trajectories as the input to the approximation, this approach does not estimate task-conditioned parameter estimates, which can lead to excessive estimation efforts. In contrast, Muratore et al. \cite{drboMuratoreEGP21} uses Bayesian Optimisation (BO) with real-world policy evaluations to estimate task-conditioned distribution statistics of simulation parameters, on which a DR agent is trained. In another attempt, Mozifian et al. \cite{mozian2020learning} uses a gradient based search method to learn the simulation parameter distribution by iteratively updating a prior distribution of simulation parameters and rolling out a DR policy trained on this distribution. On a related note, \emph{EPOpt} \cite{DBLP:conf/iclr/RajeswaranGRL17} uses a policy gradient based meta-algorithm to learn from trajectories of many models and improves the model parameter distribution using importance sampling. While providing impressive results, the common drawback of these approaches is the slow adaptation to a new environment compared to using a trained UPN, which may limit their applicability in physics tasks that need fast responses.

In this study, we incorporate uncertainty estimates of parameter estimation into a UPN based policy search workflow. Modelling uncertainty estimates has been long studied in the context of Bayesian RL methods \cite{bayesianrlsurvey}. In particular, many model-based Bayesian RL methods formulate a task as a partially observable Markov decision process (POMDP) and integrate uncertainty estimates as a belief over the state transition dynamics \cite{ross-bayesian-pomdp,ross2007bayes}. However, due to the complexity involved in modelling many belief models and integrating them into one policy outcome, generally they have not been applicable in complex RL tasks. To remedy this issue, Lee et al. \cite{lee2018bayesian} uses a neural network based belief and state encoder to combine many model beliefs when learning a policy. However, there is no straightforward mechanism to apply these methods in a BO-based policy search workflow. But we adopt a Bayesian outlook to estimate the epistemic uncertainty of simulation parameter estimates in our approach.

\section{Conclusion\label{sec:conclusion}}

In this study, we introduced a novel approach to improve the Sim2Real transfer performance while being robust to 
environment noise. Our proposed approach, \emph{UncAPS}, combines domain randomisation (DR) with task-conditioned
parameter estimation from Universal Policy Networks (UPNs) to generate robust 
policies. To calibrate the parameter ranges used for randomisation, we used 
aleatoric and epistemic uncertainty estimates calculated during the policy search
process. We empirically evaluated our approach in five MuJoCo tasks, and demonstrated
its improved transfer performance in noisy environments relative to other competing baselines.

{\small{}\bibliographystyle{IEEEtran}
\bibliography{physics}
}{\small\par}

%



\pagebreak\clearpage{}
	
\section{Appendix\label{sec:supp}}

\subsection{UncAPS Performance - Extended}

\begin{figure*}[h!t]
\centering{}
\subfloat[Ant]{\includegraphics[width=0.40\paperwidth]{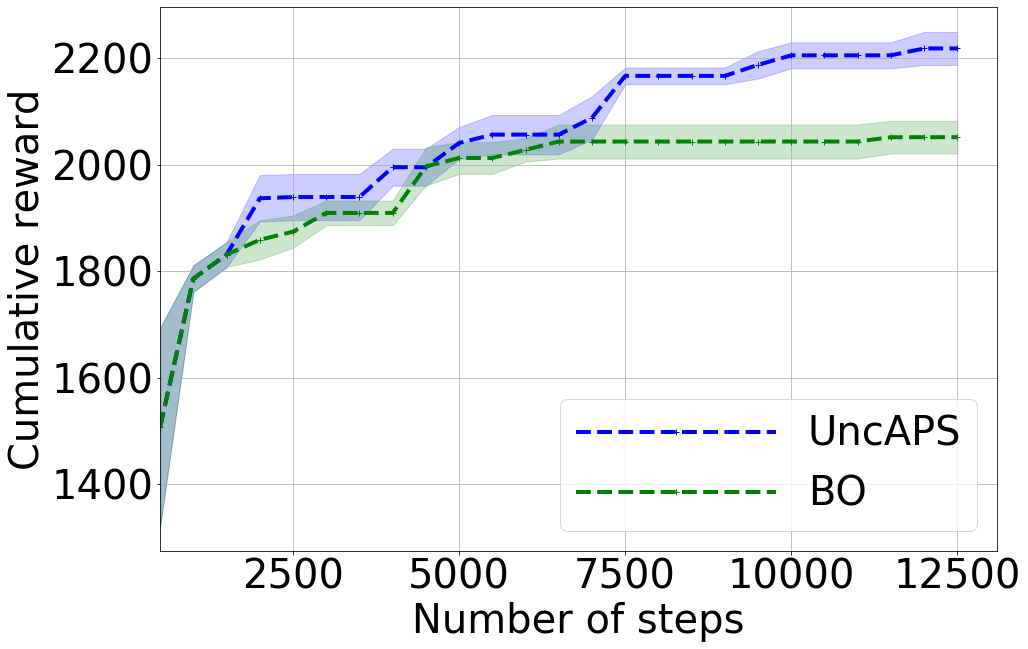}
}
\subfloat[Humanoid]{\includegraphics[width=0.40\paperwidth]{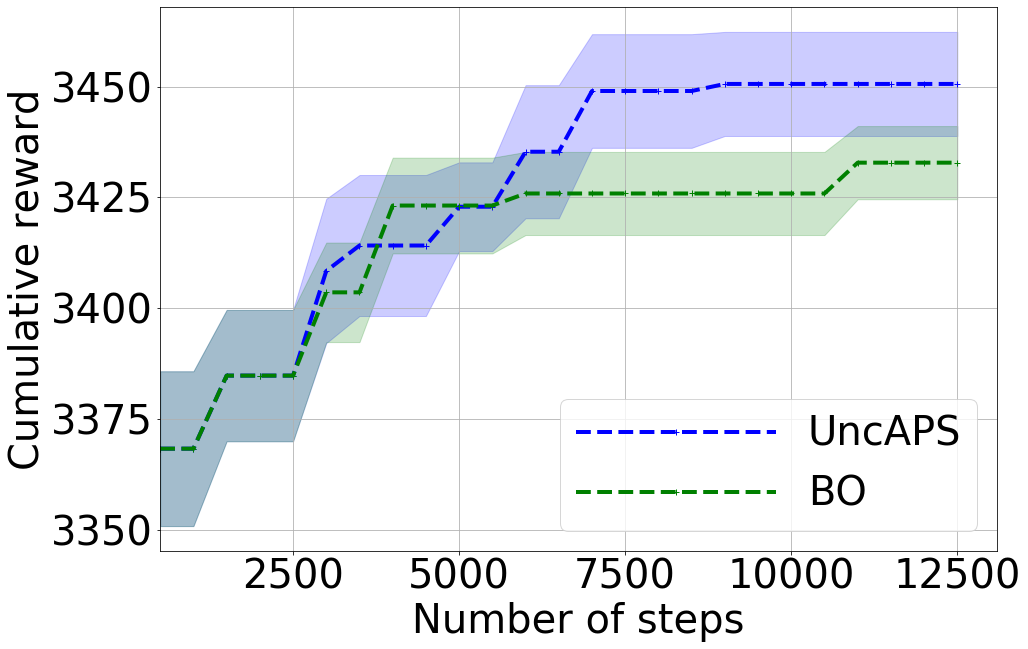}
}
\medskip{}
\subfloat[Swimmer]{\includegraphics[width=0.40\paperwidth]{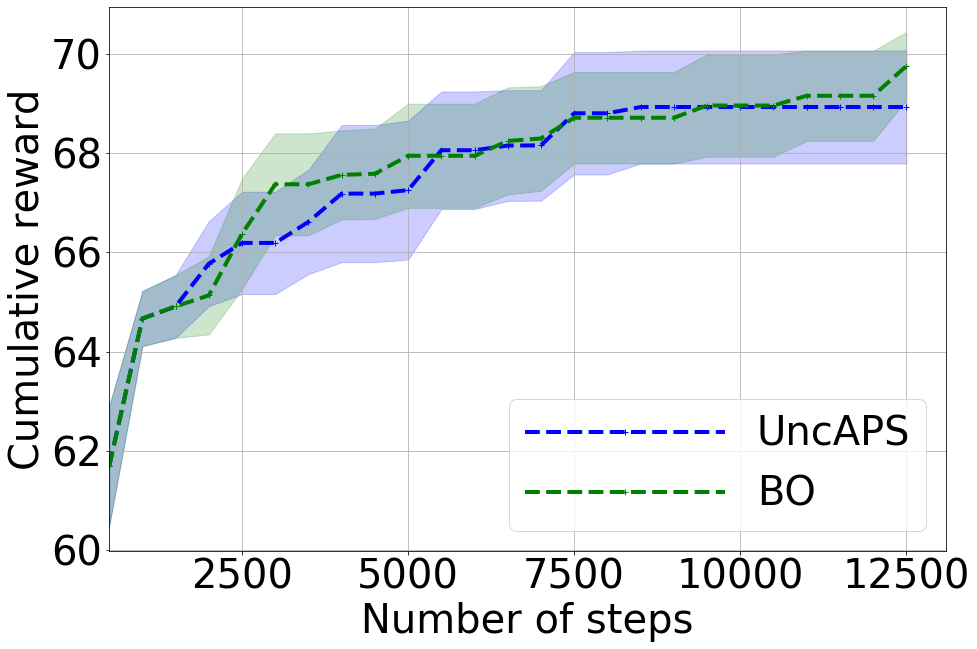}
}
\subfloat[Half-Cheetah]{\includegraphics[width=0.40\paperwidth]{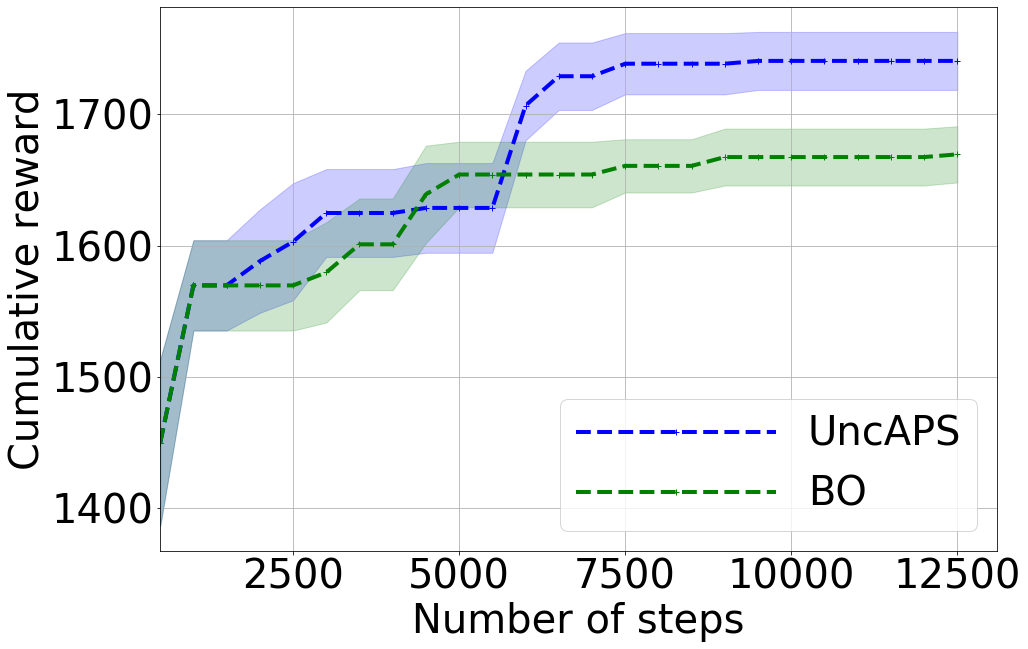}
}
\medskip{}
\subfloat[Hopper]{\includegraphics[width=0.40\paperwidth]{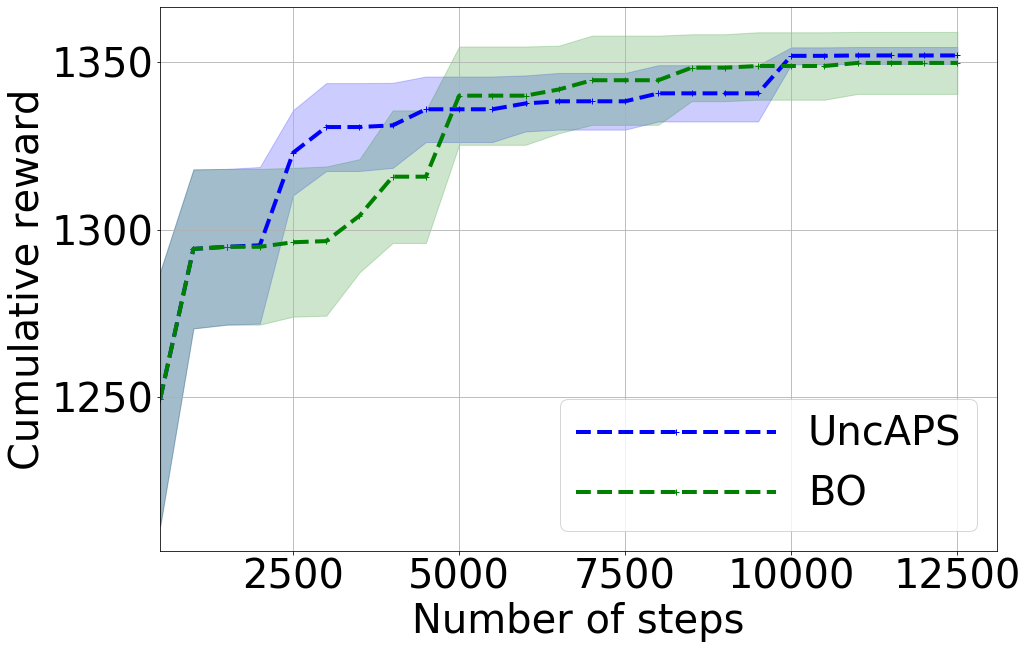}
}

\caption{\label{fig:search-results} Performance (cumulative reward) during BO policy search with \emph{UncAPS} on five MuJoCo tasks when \textbf{five} latent parameters (restitution, friction and mass of three body parts) are unknown. For all tasks a 500 step horizon is used.}
\end{figure*}

Fig. \ref{fig:search-results} shows the performance of policy search over five MuJoCo tasks with \emph{UncAPS} and standard BO, in terms of the cumulative reward over a 500 step window.

\subsection{Implementation Details \label{subsec:implementation-details}}

When building BO based policy search, we used GPy\cite{gpy2014} for GP model building 
and Emukit \cite{emukit2019} for BO process.
All BO variants reported are averaged over 5 trials, each
of which is initialised with its random
number generator seed mutually exclusively set to one of 50, 100,
150, 500 and 1000. For standard BO operations, we use RBF kernel, and Expected
improvement (EI) as the acquisition function, with LBFGS as the optimiser. For \emph{UncAPS} variants, we use Unscented Expected Improvement (UEI) as the acquisition function \cite{unscented-bo} while keeping other hyperparameters similar to standard BO.
BO processes starts with 3 random samples and proceeds to build the GP, where they
sample simulation parameters within the bounds {[}0, 1{]} as all features values are normalised.
For all tasks, we set the ground truth to a vector of 5, generated with \emph{NumPy} random.uniform(0, 1, 5)\footnote{https://numpy.org/doc/stable/reference/random/generated/numpy.random.uniform.html} and random number generator seed reset to 0.

All baselines are pre-trained for an equal number of steps
as UPN in the simulated environment. DR baseline is additionally trained for at least 12,500 real-world interactions (with upto a maximum of 500 additional interactions used due to an implementation limitation) to
match UPN's BO search budget of $25 \times 500$ interactions for all tasks. Similarly \emph{MAML} baseline uses 12,500 real-world interactions as part of its adaptation phase. In \emph{UncAPS$-$EP}, \emph{UncAPS$+$GA} and \emph{UncAPS} baselines, we set the unscented coefficient to $K=2$. When estimating the optimal simulator parameter distribution in \emph{UncAPS}, we use BFGS as the optimiser.

\end{document}